\documentclass[10pt,twocolumn,letterpaper]{article}

\usepackage{iccv}              

\usepackage{makecell}
\usepackage{tabularx}
\usepackage{multirow}
\usepackage{float}

\definecolor{iccvblue}{rgb}{0.21,0.49,0.74}
\usepackage[pagebackref,breaklinks,colorlinks,allcolors=iccvblue]{hyperref}

\def\paperID{870} 
\def\confName{ICCV}
\def\confYear{2025}

\title{EDFFDNet: Towards Accurate and Efficient Unsupervised \\ Multi-Grid Image Registration}
\author{
    Haokai Zhu$^{1,2,*}$
    \quad Bo Qu$^{2,*}$
    \quad Si-Yuan Cao$^{1,3,\dagger}$ 
    \quad Runmin Zhang$^{2}$ \\
    \quad Shujie Chen$^{4}$
    \quad Bailin Yang$^{4}$
    \quad Hui-Liang Shen$^{2}$ \\
    {\small $^{1}$Ningbo Global Innovation Center, Zhejiang University} \\
    {\small $^{2}$College of Information Science and Electronic Engineering, Zhejiang University} \\
    {\small $^{3}$NingboTech University \quad $^{4}$Zhejiang Key Laboratory of Big Data and Future E-Commerce Technology, Hangzhou, China} \\
    {\tt\small hkzhu.zju@gmail.com \{22431157, cao\_siyuan, runmin\_zhang\}@zju.edu.cn} \\
    {\tt\small \{chenshujie, ybl\}@zjgsu.edu.cn \quad shenhl@zju.edu.cn}
}

\begin{document}
\maketitle

\begingroup 
\renewcommand\thefootnote{\fnsymbol{footnote}} 
\footnotetext{$^*$ Equal Contributions. $^\dagger$ Corresponding author.}
\endgroup

\begin{abstract}
Previous deep image registration methods that employ single homography, multi-grid homography, or thin-plate spline often struggle with real scenes containing depth disparities due to their inherent limitations. To address this, we propose an Exponential-Decay Free-Form Deformation Network (EDFFDNet), which employs free-form deformation with an exponential-decay basis function. This design achieves higher efficiency and performs well in scenes with depth disparities, benefiting from its inherent locality.
We also introduce an Adaptive Sparse Motion Aggregator (ASMA), which replaces the MLP motion aggregator used in previous methods. By transforming dense interactions into sparse ones, ASMA reduces parameters and improves accuracy. Additionally, we propose a progressive correlation refinement strategy that leverages global-local correlation patterns for coarse-to-fine motion estimation, further enhancing efficiency and accuracy.
Experiments demonstrate that EDFFDNet reduces parameters, memory, and total runtime by 70.5\%, 32.6\%, and 33.7\%, respectively, while achieving a 0.5 dB PSNR gain over the state-of-the-art method. With an additional local refinement stage, EDFFDNet-2 further improves PSNR by 1.06 dB while maintaining lower computational costs. Our method also demonstrates strong generalization ability across datasets, outperforming previous deep learning methods.
\end{abstract}    

\section{Introduction}
Image registration is a fundamental task in computer vision, establishing spatial correspondence between images captured under varying conditions. It has been widely applied in various domains, including image/video stitching \cite{zaragoza2013projective, guo2016joint, nie2024eliminating}, video stabilization \cite{liu2013bundled, zhang2023minimum}, camera calibration \cite{zhang2002flexible}, HDR imaging \cite{gelfand2010multi}, and SLAM \cite{zou2012coslam}. Traditional methods typically rely on the extraction of geometric features \cite{lowe2004distinctive, rublee2011orb, leutenegger2011brisk}, followed by robust estimation strategies \cite{fischler1981random, barath2019magsac, barath2020magsac++} with outlier rejection to achieve registration. Although these methods are effective in various scenarios, they often exhibit limited robustness in low-texture scenes. 

\begin{figure}[t]
  \centering

   \includegraphics[width=1\linewidth]{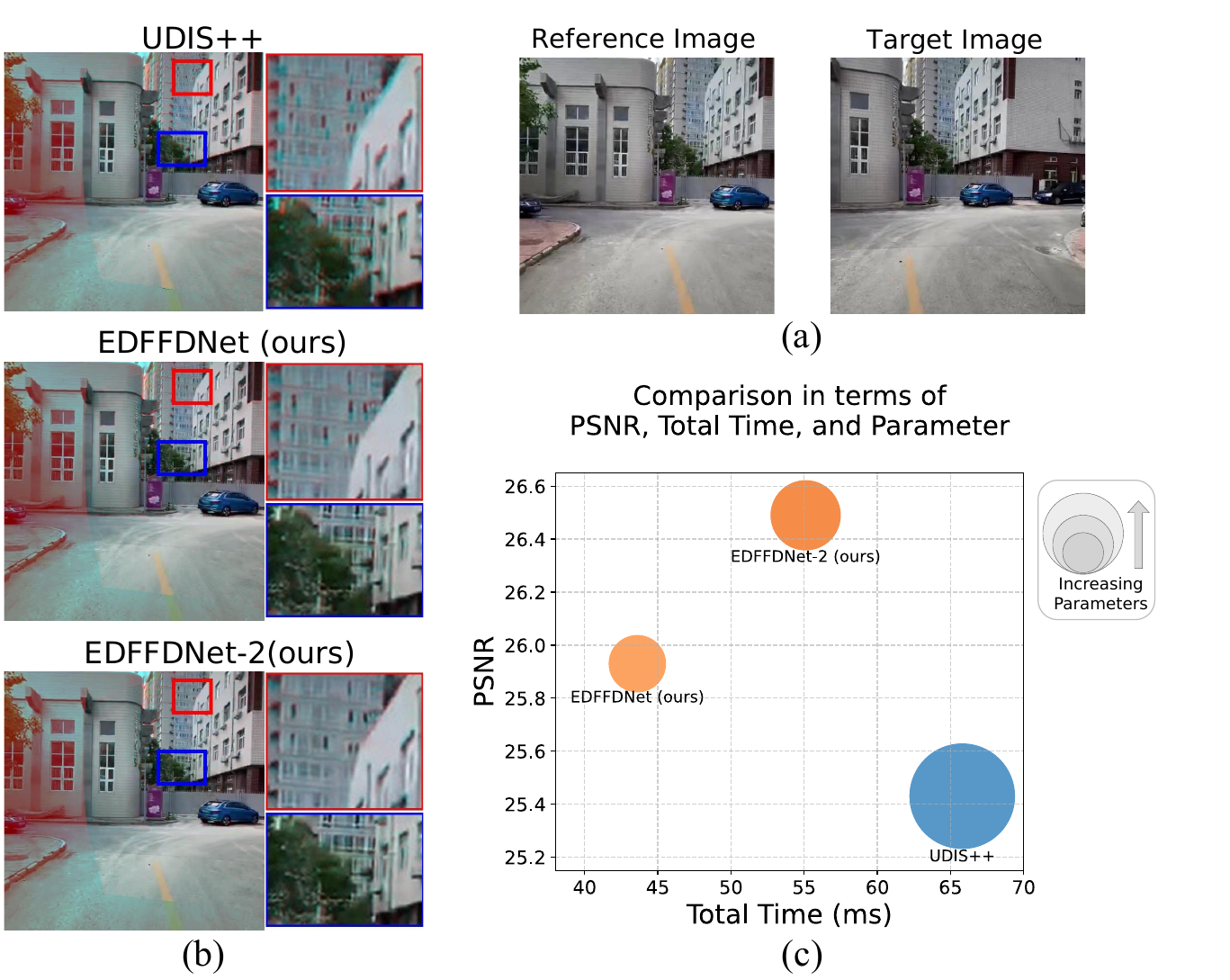}
\vspace{-4mm}
\caption{Performance comparison on the UDIS-D dataset~\cite{nie2021unsupervised}. (a) The reference and target images from a sample in the UDIS-D dataset. (b) Warp results of UDIS++ \cite{nie2023parallax}, our EDFFDNet, and EDFFDNet-2 on the sample. We combine the green and blue channels of the reference image with the red channel of the warped target image for visualization. Our method achieves better alignment in challenging regions with depth disparities (zoomed-in areas). (c) Performance comparison plot. Our method exhibits lower computational overhead while maintaining competitive performance.}
   \label{fig:head}
\vspace{-6mm}
\end{figure}

Deep learning-based methods are then proposed. Pioneer work \cite{detone2016deep} proposes a VGG-style network to estimate homography. Subsequent studies refine the network architecture \cite{erlik2017homography, zhou2019integrated, le2020deep} or integrate \cite{chang2017clkn} the IC-LK \cite{baker2004lucas} to enhance performance. Cao \etal~\cite{cao2022iterative} introduce an end-to-end homography estimation network, which is further improved through attention mechanism \cite{cao2023recurrent} and multi-scale correlation \cite{zhu2024mcnet}. However, these methods are trained in a supervised manner using synthetic datasets, which limits the generalization to real-world images. Nguyen \etal~\cite{nguyen2018unsupervised} first propose an unsupervised approach. Given that real-world images often violate the planar assumption of homography, some studies utilize predicted outlier masks \cite{le2020deep, zhang2020content} or use coplanarity constraint \cite{hong2022unsupervised} to constrain the estimation. Despite these advancements, single homography still struggles to align scenes with significant depth disparities due to its inherent limitations.

To address this challenge, more adaptive warping methods are proposed. For traditional methods, Gao \etal~\cite{gao2011constructing} propose to estimate dual-homography for two dominant planes. Zaragoza \etal~\cite{zaragoza2013projective} apply global projective warp with local deviations. Lee \etal~\cite{lee2020warping} further improves this by applying weighted homographies. Li \etal~\cite{li2017parallax} employ the thin plate spline (TPS) model \cite{bookstein1993thin} to realize flexible alignment. Learning-based methods have focused on integrating these warping models into deep networks. To stabilize the training of multi-grid homography networks, Wang \etal~\cite{wang2018deep} introduce inter-grid consistency and intra-grid regularity terms. Nie \etal~\cite{nie2021depth} propose an efficient contextual correlation layer. Since multi-grid homography fails to comprehensively tackle the limitations of homography, Wang \etal~\cite{wang2024mask} predicts pseudo plane masks and applies local homographies for each mask. Nie \etal~\cite{nie2023parallax} achieve more precise local alignment by adopting the TPS model in place of the multi-grid homography scheme. However, TPS constructs a globally smooth deformation field without direct local support, making it less effective in handling significant local deformations.

To address the aforementioned issues, we propose the Exponential-Decay Free-Form Deformation Network (EDFFDNet). The primary motivation is to better handle local deformation while reducing computational costs through more localized processing. Therefore, we adopt the Free-Form Deformation (FFD) model \cite{tustison2009directly}, which inherently offers better locality. However, conventional FFD implementations typically employ B-splines as basis functions, which provide smooth deformation but incur substantial computational overhead. To address this, we introduce an Exponential-Decay Free-Form Deformation (EDFFD) model. Experiments demonstrate that it not only improves registration accuracy but also substantially reduces computational overhead. Thanks to the better locality of EDFFD, we can achieve improved alignment performance through additional local refinements.

In the previous state-of-the-art (SOTA) method~\cite{nie2023parallax}, MLPs are used for motion aggregation. While linear layers effectively achieve global motion aggregation, their dense interactions lead to 
large computation cost, restricting deployment. Motivated by the sparse processing in depthwise separable convolution~\cite{howard2017mobilenets}, we propose the Adaptive Sparse Motion Aggregator (ASMA). ASMA transforms dense interactions into sparse ones through our proposed Group Linear Layers (GLL), followed by a simple linear layer to adaptively fuse the sparse aggregation results. Experiments demonstrate that ASMA significantly reduces parameters by about 66.6\% while preserving the accuracy.  
Furthermore, previous deep registration methods \cite{nie2021depth, nie2023parallax}) consistently use global correlation across all stages. While this offers a large receptive field for low-overlap cases, it introduces disturbance from distant regions in local refinement stages, reducing accuracy. We thus propose a progressive correlation strategy that applies global correlation for coarse motion estimation and local correlation for fine local motion estimation, realizing higher accuracy and efficiency.

Based on the above improvements, our method achieves significant performance gains compared to the SOTA method~\cite{nie2023parallax}. As shown in Fig.~\ref{fig:head}, our method demonstrates robust performance in regions with significant depth disparities with lower computational costs. Specifically, with only one local refinement, our EDFFDNet reduces the number of parameters, inference memory, and total time by 70.5\%, 32.6\%, and 33.7\%, respectively, while achieving a 0.5 dB PSNR improvement. Furthermore, with an additional local refinement, our EDFFDNet achieves a 1.06 dB PSNR gain, while still maintaining lower computational costs.

In summary, the contributions of our work are as follows:
\begin{itemize}
\item We propose EDFFDNet, an unsupervised registration framework that outperforms state-of-the-art methods in accuracy, computational efficiency, and generalization. 

\item We introduce an exponential-decay free-form deformation model that outperforms conventional TPS and B-spline FFD models in handling local deformation and enhancing computational efficiency.

\item We design an adaptive sparse motion aggregator that replaces dense interactions with sparse ones for motion aggregation, reducing parameters and improving accuracy.

\item We introduce a progressive correlation strategy that applies global-local correlation patterns for coarse-to-fine motion estimation, achieving better accuracy and efficiency.
\end{itemize}

\section{Related Work}

\subsection{Single Homography Methods}

Homography estimation methods include feature-based and learning-based approaches. Feature-based methods detect feature points~\cite{lowe2004distinctive,rublee2011orb,leutenegger2011brisk} and use robust estimators \cite{fischler1981random,barath2019magsac, barath2020magsac++} to solve homography, but struggle in low-texture regions.
Learning-based methods learn robust representations effectively. DeTone \etal~\cite{detone2016deep} first proposed a VGG-style \cite{simonyan2014very} deep homography network. Recent supervised advances include end-to-end estimators \cite{cao2022iterative}, attention mechanisms \cite{cao2023recurrent}, and multi-scale correlation searching \cite{zhu2024mcnet} for higher accuracy. However, supervised methods trained on synthetic data generalize poorly to real images. Nguyen \etal~\cite{nguyen2018unsupervised} first trained a homography network on real data unsupervised, while Jiang \etal~\cite{jiang2023supervised} generated realistic datasets from real images. To improve accuracy, Zhang \etal~\cite{zhang2020content} and Le \etal~\cite{le2020deep} used predicted masks to eliminate moving objects, and Hong \etal~\cite{hong2022unsupervised} introduced coplanarity constraints for dominant plane registration. However, single-homography methods fundamentally lack representational capacity for scenes with depth disparities.

\subsection{Adapative Warping Methods} 
Various approaches have been proposed to address challenges caused by depth disparities. Gao \etal~\cite{gao2011constructing} introduced a dual-homography warping model for two dominant planes. Zaragoza \etal~\cite{zaragoza2013projective} proposed APAP, which combines global homography with local homographies. Lee \etal~\cite{lee2020warping} improved this by partitioning images into superpixels and estimating weighted homographies. Li \etal~\cite{li2017parallax} employed the thin plate spline (TPS) model to handle parallax in image warping. However, these feature-based methods often fail in low-texture scenarios.
For learning-based approaches, Wang \etal~\cite{wang2018deep} introduced inter-grid consistency and intra-grid regularity terms to stabilize the training of multi-grid homography networks. Nie \etal~\cite{nie2021depth} proposed a contextual correlation layer to improve efficiency. However, multi-grid homography does not fully address the restriction of homography. Wang \etal~\cite{wang2024mask} then proposed multi-plane homography estimation, which and applies local homographies for each predicted pseudo-plane. Nie \etal~\cite{nie2023parallax} replaced the multi-grid homography scheme with TPS to achieve more precise and efficient local alignment. However, TPS constructs a globally smooth deformation field without local support, which limits its performance in scenarios requiring more localized deformations.

\begin{figure*}[t]
  \centering
\vspace{-1mm}
   \includegraphics[width=1\linewidth]{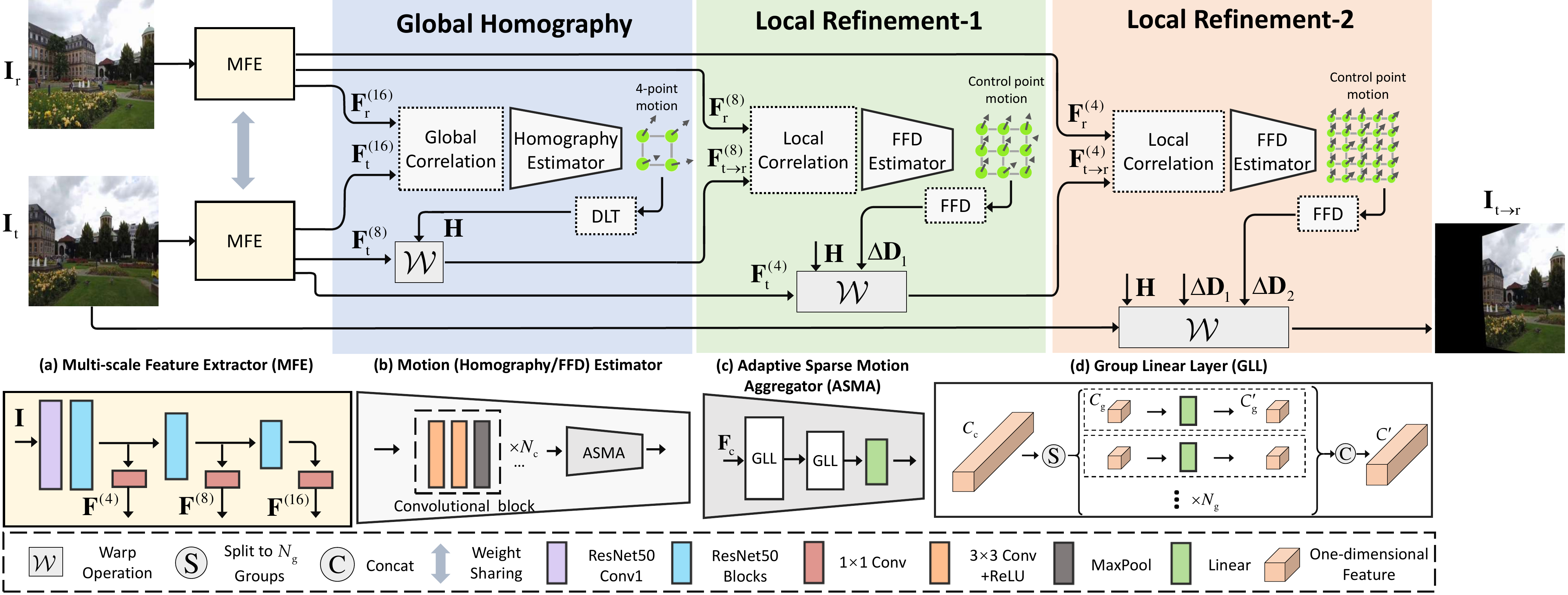}
\vspace{-7mm}
   \caption{An overview of the proposed Exponential-Decay Free-Form Deformation Network (EDFFDNet). (a) Multi-scale Feature Extractor (MFE). (b) Motion (Homography/FFD) Estimator. (c) Adaptive Sparse Motion Aggregator (ASMA). (d) Group Linear Layer (GLL).}
   \label{fig:EDFFD}
\vspace{-6mm}
\end{figure*}

\section{Method}

\subsection{Network Architecture}
As illustrated in Fig. \ref{fig:EDFFD}, the proposed Exponential-Decay Free-Form Deformation Network (EDFFDNet) consists of three main modules: multi-scale feature extractor (MFE), global homography estimation module, and local refinement module. Initially, multi-scale features are extracted from the input target image $\mathbf{I}_\mathrm{t}$ and reference image $\mathbf{I}_\mathrm{r}$. The global homography module then estimates a global homography $\mathbf{H}$, which serves as an initial alignment for subsequent local refinements. Taking the number of local refinement stages $N_{\mathrm{s}}=2$ as an example, in each stage $i$, multi-scale features are utilized to compute local correlations and estimate control point motions. These motions are then used to generate residual displacements $\Delta\mathbf{D}_i$ based on the proposed Exponential-Decay Free-Form Deformation (EDFFD) model. Finally, the residual displacements \(\Delta\mathbf{D}_2\) from the last refinement stage are combined with the global homography \(\mathbf{H}\) and the residual displacements \(\Delta\mathbf{D}_1\) from the previous local refinement stage. This combination is used to align the target image \(\mathbf{I}_\mathrm{t}\) with the reference image \(\mathbf{I}_\mathrm{r}\), resulting in the aligned image \(\mathbf{I}_{\mathrm{t}\xrightarrow{}\mathrm{r}}\).

\textbf{Feature Extraction.} 
As illustrated in Fig. \ref{fig:EDFFD}a, the target image $\mathbf{I}_\mathrm{t}$ and the reference image $\mathbf{I}_\mathrm{r}$ with size $H\times W$ are fed into MFE to extract multi-scale feature maps $\mathbf{F}_{\mathrm{t}}^{(d)}, \mathbf{F}_{\mathrm{r}}^{(d)}$, where $d \in \{4, 8, 16\}$ denotes the downsample scale. MFE mainly consists of  ResNet50 blocks \cite{he2016deep}, and a $1\times1$ convolution is applied to the feature output at each resolution. 

\textbf{Motion Estimation.} 
As shown in Fig. \ref{fig:EDFFD}b, the correlation result from the correlation computation is fed into the motion estimator, which estimates motion parameters. Specifically, it estimates 4-point motion for homography and control point motion for free-form deformation (FFD). The motion estimator consists of \(N_c\) convolutional blocks and an Adaptive Sparse Motion Aggregator (ASMA). Each convolutional block consists of $3\times 3$ convolutions, ReLU, and max-pooling. Convolutional blocks are used to extract the latent motion feature, while the ASMA transforms the latent motion feature into motion parameters. The details of the ASMA will be discussed in the corresponding section. 

\subsection{Exponential-Decay Free-Form Deformation}
Restricted by the planar assumption, homography transformation often exhibits inaccurate alignment in non-planar scenes due to its inherent limitation. The multi-grid homography scheme is proposed \cite{zaragoza2013projective} but lacks efficient parallel acceleration capabilities for deep learning \cite{nie2021depth}. The thin plate spline (TPS) transformation model \cite{bookstein1993thin} is then employed to achieve efficient and flexible deformation \cite{nie2023parallax}, but it inherently struggles to handle substantial local deformation. In contrast, B-spline free-form deformation (B-spline FFD) \cite{tustison2009directly} provides better locality essentially compared to TPS. However, the cubic B-spline basis computation incurs significant computational overhead. To address this, we propose the Exponential-Decay Free-Form Deformation (EDFFD) method, which achieves more efficient local deformation through an improved basis function.

Taking the local refinement stage $i$ as an example, for B-spline FFD, let $P = \{\mathbf{p}_{m,n} \mid 0 \leq m \leq M_i, 0 \leq n \leq N_i\}$ denote the set of control points uniformly distributed on the image, forming a mesh grid of size $M_i \times N_i$. The motion of the control point $\mathbf{p}_{m,n}$ estimated by the network is denoted as $\Delta\mathbf{p}_{m,n}$. The deformation of a point $\mathbf{x} = (x_1, x_2)$ in the plane is directly given by:
\begin{equation}
\mathbf{x}' = \mathbf{x} + \sum_{m=0}^{M_i} \sum_{n=0}^{N_i} \Delta\mathbf{p}_{m,n} \Phi\left((\mathbf{x} - \mathbf{p}_{m,n})/\eta\right),
\end{equation}
where $\eta$ represents the grid spacing, $\mathbf{x}'$ is the deformed position, and $\Phi(\cdot)$ denotes the basis product, defined as:
\begin{equation}
\Phi(\mathbf{u}) = \beta^3(u_1) \beta^3(u_2),
\end{equation}
where $\mathbf{u}=(u_1,u_2)$ denotes the input vector,  $\beta^3(\cdot)$ is the cubic B-spline function obtained by three times convolution of the zeroth-order B-spline function \cite{unser1999splines} (\textit{For more details, please refer to Section A.1 of supplementary material.}), formulated as:
\begin{align}
\beta^3(u) = \begin{cases}
\frac{2}{3}-|u|^2+\frac{|u|^3}{2}, & 0 \leq|u| \leq 1 \\
\frac{(2-|u|)^3}{6}, & 1 \leq |u| < 2, \\
0, & 2\leq|u|.
\end{cases}
\end{align}
Cubic B-spline has two principal characteristics that make them effective for deformation tasks \cite{tustison2009directly}: 1) Locality, where each control point influences a limited neighborhood, with its influence diminishing over distance, enabling localized shape manipulation, and 2) $C^2$ continuity, ensuring smooth deformation fields for natural image alignment. However, the computation presents three issues: 1) High polynomial computation cost, 2) Basis product that requires separate basis computation in two dimensions, and 3) Piecewise computation that impedes GPU parallelism.
To overcome these limitations while preserving beneficial properties, we propose EDFFD as an alternative to B-spline FFD, defined as: 
\begin{align}
\mathbf{x}' = \mathbf{x} + \sum_{m=0}^{M_i} \sum_{n=0}^{N_i} \Delta\mathbf{p}_{m,n} \mathrm{exp}\left(-r_{m,n}/(\theta\eta)\right),
\end{align}
where $\theta$ is a factor controlling the decay rate of influence, $r_{m,n}$ represents the Euclidean distance, defined as: 
\begin{align}
r_{m,n} = \| \mathbf{x}-\mathbf{p}_{m,n} \|_2.
\end{align}
This model is motivated by four key principles:
\begin{itemize}
    \item Simplified Influence Metric: Instead of combining control point influences in two dimensions via basis products, we directly use Euclidean distance, eliminating expensive separate basis computation.
    \item Computational Efficiency: The exponential function provides $C^\infty$ smoothness with lower computational overhead than cubic polynomials. Modern GPU architectures further accelerate its implementation through hardware-optimized transcendental units.
    \item Parallel Compatibility: The non-piecewise nature of the exponential function enables fully parallel evaluation across spatial domains, contrasting with B-spline's conditional branching that hinders GPU utilization.
    \item Locality Preservation: The exponential function naturally decays significantly with distance, ensuring that the influence of each control point remains localized.
\end{itemize}

\subsection{Adaptive Sparse Motion Aggregation}
\begin{figure}[t]
  \centering
   \includegraphics[width=1\linewidth]{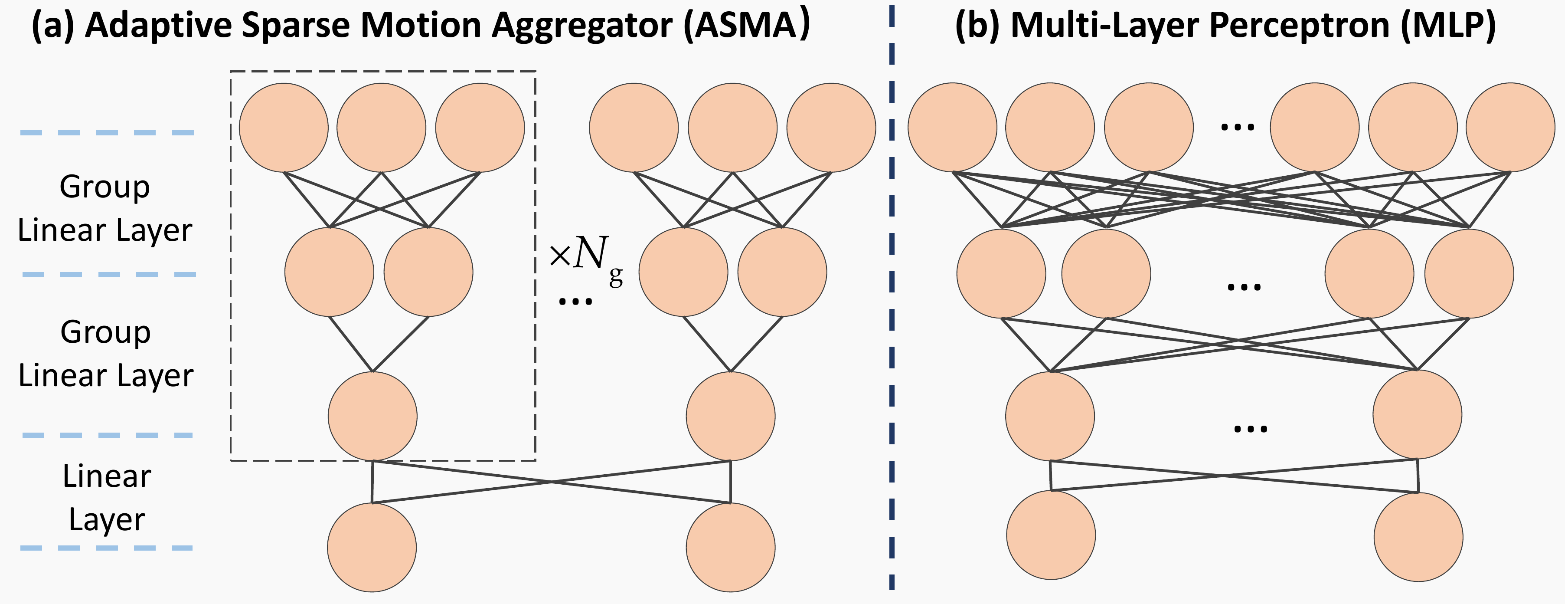}
\vspace{-7mm}
\caption{Comparison of neuron interactions between our Adaptive Sparse Motion Aggregator (ASMA) and a Multi-Layer Perceptron (MLP). ASMA achieves efficient motion aggregation by sparsely processing grouped features.}
   \label{fig:ASMA_fig}
\vspace{-7mm}
\end{figure}

Previous work \cite{nie2023parallax} utilizes convolutional blocks and MLP to construct the motion estimator. Although linear layers are computationally efficient and perform well in motion aggregation, they suffer from high parameter counts, limiting their practical deployment. To address this limitation, we propose an Adaptive Sparse Motion Aggregator (ASMA) inspired by the sparse feature processing in depthwise separable convolutions~\cite{howard2017mobilenets}, which reduces computational overhead while preserving accuracy. ASMA comprises two Group Linear Layers (GLL) and a single linear layer. The group linear layers transform dense interactions into sparse ones, while the linear layer adaptively fuses the sparse motion aggregation results. Fig. \ref{fig:EDFFD}c and Fig.~\ref{fig:EDFFD}d illustrate the structure of ASMA, and Fig.~\ref{fig:ASMA_fig} highlights the differences between ASMA and traditional MLP.

Let $\mathbf{F}_{\mathrm{c}} \in \mathbb{R}^{C_{\mathrm{c}}}$ denote the flattened latent motion feature extracted from $N_\mathrm{c}$ convolutional blocks. This feature is first fed into GLL, which partitions $\mathbf{F}_{\mathrm{c}}$ into $N_{\mathrm{g}}$ groups to form the group feature $\mathbf{F}_{\mathrm{g},k}\in \mathbb{R}^{C_{\mathrm{g}}},k = 1, \cdots, N_{\mathrm{g}}$, 
where $C_{\mathrm{g}} = C_{\mathrm{c}} / N_{\mathrm{g}}$ assuming $C_{\mathrm{c}}$ is divisible by $N_{\mathrm{g}}$. Each group feature $\mathbf{F}_{\mathrm{g},k}$ is then passed through an independent linear layer to produce output group feature $\mathbf{F}'_{\mathrm{g},k} \in \mathbb{R}^{C'_{\mathrm{g}}}$ as follows:
\begin{align}
\mathbf{F}'_{\mathrm{g},k}=\mathbf{W}_{k}(\mathbf{F}_{\mathrm{g},k})+\mathbf{b}_{k},
\end{align}
where $\mathbf{W}_{k}$ and $\mathbf{b}_{k}$ are the weight and bias of the $k$-th linear layer. All the output group features are concatenated and then passed through the ReLU activation $\sigma$ to obtain the output feature $\mathbf{F}'\in \mathbb{R}^{C'}$ of the current GLL as follows:
\begin{align}
\mathbf{F}'=\sigma(\text{Concat}(\mathbf{F}'_{\mathrm{g},1},\cdots,\mathbf{F}'_{\mathrm{g},N_{\mathrm{g}}})).
\end{align}
GLL effectively conducts feature interactions within each group by transforming the dense connections of a traditional linear layer into sparse ones essentially. 
After passing through two GLLs, the sparse aggregation motion results are adaptively fused by a simple linear layer. This fusion serves as a global interaction mechanism and is responsible for outputting the motion parameters.

\subsection{Progressive Correlation Strategy}
Low-overlap cases in real-world images require global correlation for wide search ranges, but this is computationally expensive and disturbs local refinement. While \cite{nie2021depth} proposes efficient global correlation, successive computations still incur significant overhead \cite{nie2023parallax}. Inspired by the observation that required search ranges decrease as estimation accuracy improves, we propose a progressive correlation strategy: using global correlation in the global homography stage for broad search, then transitioning to local correlation in refinement stages for improved efficiency and accuracy.

For global correlation computation, we adopt the patch-to-patch correlation approach \cite{nie2021depth}. For the target feature $\mathbf{F}_{\mathrm{t}}^{(d)}$, dense patches of size $K \times K$ with stride 1 are extracted as convolutional filters. These filters are applied to the reference feature $\mathbf{F}_{\mathrm{r}}^{(d)}$ to form a global correlation volume $\mathbf{C}^{\mathrm{g}}$ with shape $H^{(d)} \times W^{(d)} \times H^{(d)}W^{(d)}$. Each value in the column can be formulated as: 
\begin{equation}
\mathbf{C}^{\mathrm{g}}_{(x_r,y_r,x_t,y_t)} = 
\sum_{i,j=-\lfloor{\frac{K}{2}}\rfloor}^{\lfloor{\frac{K}{2}}\rfloor} 
 \frac{ \left\langle \mathbf{F}_{\mathrm{r},(x_r+i, y_r+j)}^{(d)}, \mathbf{F}_{\mathrm{t},(x_t+i, y_t+j)}^{(d)} \right\rangle }
 {\left\| \mathbf{F}_{\mathrm{r},(x_r+i, y_r+j)}^{(d)} \right\| \left\| \mathbf{F}_{\mathrm{t},(x_t+i, y_t+j)}^{(d)} \right\|}. \\
\end{equation}
Each position in $\mathbf{C}^{\mathrm{g}}$ can be regarded as a vector of length $H^{(d)} W^{(d)}$. The vectors are first scaled by a constant $\alpha$ to increase the in-class distance and then passed through a softmax to obtain matching probabilities. The positions with the highest probabilities are thereby identified, from which we derive the feature flow $\mathbf{V}$ with shape $H^{(d)} \times W^{(d)} \times 2$ as the global correlation result.

For local correlation computation, we adopt the local correlation method utilized in \cite{zhu2024mcnet} that directly computing the correlation between the feature at position \(\mathbf{p}\) in the reference feature \(\mathbf{F}_{\mathrm{r}}^{(d)}\) and features within the local area in the target feature \(\mathbf{F}_{\mathrm{t}}^{(d)}\). The local correlation is formulated as:
\begin{align}
\mathbf{C}^{\mathrm{l}}(\mathbf{p}, \mathbf{p}^{\prime}) = \mathbf{F}_{\mathrm{r}}^{(d)}(\mathbf{p})^{\top} \mathbf{F}_{\mathrm{t}}^{(d)}(\mathcal{A}(\mathbf{p}^{\prime}, r)),
\end{align}
where $\mathcal{A}(\mathbf{p}^{\prime}, r))$ represents the sampling local area centered at \(\mathbf{p}^{\prime}\) with radius \(r\), $\mathbf{C}^{\mathrm{l}}$ is the local correlation result with shape $H^{(d)} \times W^{(d)} \times (2r+1)^2$.

\subsection{Optimization}
To achieve both precise content alignment and natural shape preservation in the image registration results, we optimize the network using two terms. The content alignment term aligns the input images according to their content, while the shape preservation term prevents unnatural distortions.

\textbf{Content Alignment.} Consider a reference image $\mathbf{I}_\mathrm{r}$ and a target image $\mathbf{I}_\mathrm{t}$. Let $\mathcal{W}(\cdot, \cdot)$ represent the warping operation, and $\mathbf{J}$ denote an all-one matrix with the same resolution as the image. We define the content alignment loss as follows:
\begin{align}
    \mathcal{L}_{\mathrm{content}} &= \lambda_0 \|\mathbf{I}_\mathrm{r}\cdot \mathcal{W}(\mathbf{J}, \mathcal{H})-\mathcal{W}(\mathbf{I}_\mathrm{t},\mathcal{H})\|_1 \\ \notag
    &\quad + \lambda_0 \|\mathbf{I}_\mathrm{t}\cdot \mathcal{W}(\mathbf{J}, \mathcal{H}^{-1})-\mathcal{W}(\mathbf{I}_\mathrm{r},\mathcal{H}^{-1})\|_1 \\ \notag
    &\quad + \sum_{i=1}^{N_\mathrm{s}} \lambda_i \| \mathbf{I}_\mathrm{r}\cdot \mathcal{W}(\mathbf{J}, \mathcal{FFD}_i)-\mathcal{W}(\mathbf{I}_\mathrm{t},\mathcal{FFD}_i)\|_1 ,
\end{align}
where $\lambda_0$ and $\lambda_i$ are weights, $\mathcal{H}$ and $\mathcal{FFD}_i$ denote the warping parameters of global homography and free-form deformation at the local refinement stage $i$, respectively.

\textbf{Shape Preservation.} As in \cite{nie2023parallax}, we preserve the shape by leveraging the inter-grid constraint $\ell_{\mathrm{inter}}$ and intra-grid constraint $\ell_{\mathrm{inter}}$ based on the grid edge $\vec{e}$ as follows:
\begin{equation}
    \mathcal{L}_{\mathrm{shape}}=\sum_{i=1}^{N_\mathrm{s}}(\ell_{\mathrm{intra},i}+\ell_{\mathrm{inter},i}).
\end{equation}
Let $\{\vec{e}_{\mathrm{h}, i}\}$ and $\{\vec{e}_{\mathrm{v}, i}\}$ represent the sets of horizontal and vertical edges of local refinement stage $i$, respectively. The intra-grid constraint is defined as follows:
\begin{align}
\ell_{\mathrm{intra},i}= & \frac{1}{(M_i+1)\times N_i}(\sum_{\{\vec{e}_{\mathrm{h},i}\}}\sigma(\langle\vec{e},\vec{u}\rangle-\frac{2W}{N_i})+ \\ 
 & \frac{1}{M_i\times (N_i+1)}\sum_{\{\vec{e}_{\mathrm{v},i}\}}\sigma(\langle\vec{e},\vec{v}\rangle-\frac{2H}{M_i})), \notag
\end{align}
where $\sigma$ denotes the ReLU activation function, $\vec{u}$ and $\vec{v}$ are unit vectors along the $x$ and $y$ directions, respectively.
For the non-overlapping region, the consecutive edges are encouraged to be colinear to preserve the structures. The inter-grid constraint is defined as follows:
\begin{equation}
    \ell_{\mathrm{inter},i}=\frac{1}{E}\sum_{\{\vec{e}_{\mathrm{c1},i},\vec{e}_{\mathrm{c2},i}\}}Q_{c1,c2}\cdot(1-\frac{\langle\vec{e}_{\mathrm{c1}},\vec{e}_{\mathrm{c2}}\rangle}{\parallel\vec{e}_{\mathrm{c1}}\parallel\cdot\parallel\vec{e}_{\mathrm{c2}}\parallel}),
\end{equation}
where $\{\vec{e}_{\mathrm{c1},i},\vec{e}_{\mathrm{c2},i}\}$ represents the set of edge pairs that are consecutive in either the horizontal or vertical direction of the local refinement stage $i$, $E$ denotes the number of edge pairs, and $Q_{\mathrm{c1},\mathrm{c2}}$ is 1 for edge pairs in non-overlapping regions and 0 otherwise. The entire loss function is defined as follows:
\begin{equation}
    \mathcal{L}=\mathcal{L}_{\mathrm{content}}+\omega\mathcal{L}_{\mathrm{shape}},
\end{equation}
where $\omega$ is the weight to balance the loss.

\section{Experiments}

\subsection{Implementation Details and Datasets}
Network training proceeds in two phases: first, the global homography module is trained for 10 epochs, then training extends to 100 epochs with joint optimization of both global and local modules. For global correlation, patch size $K=3$ and $\alpha=10$; for local correlation, radius $r=4$. Loss weights are $\lambda_0=1$, $\lambda_1=1.3$, $\lambda_2=1.7$, and $\omega=10$. Training uses batch size 4 and Adam optimizer with learning rate $10^{-4}$ on PyTorch. Experiments run on NVIDIA RTX 4090 GPU, Intel Xeon Platinum 8352V CPU @ 2.10GHz, and 512GB RAM.

We evaluate our network on the UDIS-D dataset~\cite{nie2021unsupervised}, which contains 10,440 training and 1,106 testing image pairs with diverse overlap ratios and scenes. Following prior work~\cite{nie2021unsupervised, nie2021depth, nie2023parallax}, we assess performance using PSNR and SSIM within overlapping regions. We also evaluate zero-shot performance on other datasets \cite{zaragoza2013projective,gao2011constructing,dai2017scannet,schops2017multi}.

\subsection{Ablation Studies}
We conduct ablation studies on the deformation model, motion aggregation, and correlation computation. Unless specified, the default configuration uses EDFFD with $N_\mathrm{s}=1,M_1=12, N_1=12, \theta=0.75$, ASMA with $N_\mathrm{g}=8$, and progressive correlation. \textbf{Inference time} is the time for the network to estimate motion parameters from input image pairs, \textbf{warp time} is the time for the deformation model to output aligned images from motion parameters, and \textbf{total time} is their sum.

\subsubsection{Deformation Model}
As shown in Table~\ref{tab:deformation_models}, we compare the Thin Plate Spline (TPS), B-spline Free-Form Deformation (B-spline FFD), and our proposed EDFFD under identical network settings. The results demonstrate that B-Spline FFD achieves more precise registration than the TPS model due to its better locality. However, B-spline FFD incurs a significantly larger computational overhead. Our proposed EDFFD achieves nearly the same registration performance as B-spline FFD, while significantly reducing the warp time and inference memory by 47.3\% and 34.0\%, respectively. This highlights the effectiveness of our improvements in the basis function. Additionally, our EDFFD further reduces warp time by 32.4\% compared with the TPS model, due to its more simplified computation. This further indicates the efficiency and accuracy of our proposed EDFFD.

\begin{table}[h]
\renewcommand\arraystretch{1}
\renewcommand\tabcolsep{5pt}{}
\centering
\footnotesize 
\vspace{-3mm}
\caption{Ablation on deformation models.}
\vspace{-3mm}
\begin{tabular} {ccccc}
\toprule
\makecell{Model}& PSNR & SSIM & \makecell{Warp time (ms)}  & \makecell{Memory (GB)} \\
\hline
TPS & 25.49 & 0.838 & 30.5 & 3.3 \\
\makecell{B-spline FFD} & 25.95 & 0.850 & 39.1 & 4.7 \\
\makecell{EDFFD} & 25.93 & 0.852 & 20.6 & 3.1 \\
\bottomrule
\vspace{-6mm}
\end{tabular}
\label{tab:deformation_models}
\end{table}

\begin{table}[h]
\renewcommand\arraystretch{1}
\renewcommand\tabcolsep{7pt}
\centering
\footnotesize 
\vspace{-5mm}
\caption{Ablation on influence factor $\theta$ settings.}
\vspace{-3mm}
\begin{tabular}{ccccccc}
\toprule
$\theta$ & 0.25 & 0.50 & 0.75 & 1.00 & 1.25 & 1.50 \\
\hline
PSNR & 25.30 & 25.93 & 25.93 & 25.91 & 25.85 & 25.81 \\
SSIM & 0.834 & 0.852 & 0.852 & 0.851 & 0.849 & 0.848\\
\bottomrule
\vspace{-6mm}
\end{tabular}
\label{tab:theta}
\end{table}

\begin{table}[h]
\renewcommand\arraystretch{1}
\renewcommand\tabcolsep{3pt}{}
\centering
\footnotesize
\vspace{-5mm}
\caption{Ablation on local refinement settings.}
\vspace{-3mm} 
\begin{tabular} {ccccc}
\toprule
        \makecell{Stage1 Grid size} & \makecell{Stage2 Grid size} & PSNR & SSIM & \makecell{Warp time (ms)}\\ \hline
        12$\times$12 & N/A & 25.93 & 0.852 & 20.6\\ 
        12$\times$12 & 12$\times$12 & 26.43 & 0.866 & 22.1 \\ 
        12$\times$12 & 18$\times$18 & 26.49 & 0.868 & 23.8\\ 
        12$\times$12 & 24$\times$24 & 26.55 & 0.869 & 25.2\\ 
\bottomrule
\vspace{-7mm}
\end{tabular}
\label{tab:iteration}
\end{table}

\textbf{Impact of locality.} 
By design, smaller $\theta$ increases influence range, enhancing smoothness but weakening locality, and vice versa. Table~\ref{tab:theta} shows a noticeable drop at $\theta = 0.25$, while larger $\theta$ limits influence range, causing poor smoothness and reduced accuracy. These results indicate that our model's improved performance primarily benefits from well-balanced locality scope.

\textbf{Additional Local Refinement.} As shown in Table~\ref{tab:iteration}, we achieve significant accuracy improvements through an additional local refinement stage with denser grid settings, while incurring minor increases in warp time. This highlights the efficiency and potential performance of our EDFFD.

\subsubsection{Motion Aggregation}
We investigate different motion aggregation methods in Table~\ref{tab:asma}. ASMA outperforms MLP across all settings, achieving better accuracy while reducing parameters by 66.6\% for $N_{\mathrm{g}}=8$. More importantly, when MLP's hidden dimension is reduced by 4 times to match ASMA's parameter count, ASMA still maintains superior performance. This demonstrates the effectiveness of our ASMA.

\begin{table}[h]
\renewcommand\arraystretch{1}
\renewcommand\tabcolsep{2pt}{}
\centering
\footnotesize
\vspace{-4mm}
\caption{Ablation on motion aggregation methods.}
\vspace{-3mm}
\begin{tabular} {cccccc}
\toprule
        \makecell{Motion\\aggregation} & \makecell{Hidden dimension\\ reduction ratio} & \makecell{$N_{\mathrm{g}}$} & \makecell{Parameters (M)} & PSNR & SSIM  \\ \hline
        MLP & 1 & N/A & 68.9  & 25.87  & 0.850 \\ 
        ASMA & 1 & 4 & 24.3  & 25.91  & 0.851  \\ 
        ASMA & 1 & 8 & 23.0  & 25.93  & 0.852  \\ 
        ASMA & 1 & 16 & 22.3  & 25.90  & 0.851 \\ 
\hline
        MLP & 4 & N/A & 27.1  & 25.76  & 0.845 \\ 
        ASMA & 4 & 8 & 17.3  & 25.89  & 0.850 \\ 
\bottomrule
\end{tabular}
\label{tab:asma}
\end{table}

\begin{table}[h]
\renewcommand\arraystretch{1}
\renewcommand\tabcolsep{4pt}{}
\centering
\footnotesize
\vspace{-8mm}
\caption{Ablation on correlation settings.}
\vspace{-3mm}
\begin{tabular} {cccccc}
\toprule
        \makecell{Homography\\correlation} & \makecell{FFD\\correlation} & PSNR & SSIM & \makecell{Inference time (ms)} &  \\ \hline
        Global & Global & 25.54 & 0.843 & 32.8  \\ 
        Local & Local & N/A & N/A & N/A \\
        Global & Local & 25.93 & 0.852 & 23.0  \\ 
\bottomrule
\vspace{-8mm}
\end{tabular}
\label{tab:correlation}
\end{table}

\begin{table*}[t]
    \centering
    \small
    \caption{Quantitative comparison of warp on UDIS-D dataset~\cite{nie2021unsupervised} The best is marked in red and the second best is in blue.}
    \vspace{-3mm}
    \label{tab:warp_quality_comparison}
	\begin{tabular}{lcccc|cccc}
		\toprule
		& \multicolumn{4}{c|}{PSNR $\uparrow$} & \multicolumn{4}{c}{SSIM $\uparrow$} \\
		\cmidrule(lr){2-5} \cmidrule(lr){6-9}
		& Easy & Moderate & Hard & Average & Easy & Moderate & Hard & Average \\
		\midrule
		$I_{3\times3}$ & 15.87 & 12.76 & 10.68 & 12.86 & 0.530 & 0.286 & 0.146 & 0.303 \\
		SIFT~\cite{lowe2004distinctive}+RANSAC~\cite{fischler1981random} & 27.75 & 24.03 & 18.46 & 22.98 & 0.906 & 0.828 & 0.627 & 0.758\\
		APAP~\cite{zaragoza2013projective} & 27.01 & 23.39 & 19.54 & 23.00 & 0.885 & 0.802 & 0.663 & 0.773 \\
		ELA~\cite{li2017parallax} & 29.87 & 25.51 & 19.68 & 24.47 & 0.924 & 0.865 & 0.713 & 0.821 \\
		SPW~\cite{liao2019single} & 27.29 & 22.83 & 16.94 & 21.80 & 0.888 & 0.764 & 0.504 & 0.696  \\
		LPC~\cite{jia2021leveraging} & 27.04 & 22.72 & 19.34 & 22.65 & 0.879 & 0.768 & 0.610 & 0.738 \\
		UDIS~\cite{nie2021unsupervised} & 27.84 & 23.95 & 20.70 & 23.80 & 0.902 & 0.830 & 0.685 & 0.793 \\
        MGDH~\cite{nie2021depth} & 29.52 & 25.24 & 21.20 & 24.89 & 0.923 & 0.859 & 0.708 & 0.817\\
        UDIS++~\cite{nie2023parallax} & 30.19 & 25.84 & 21.57 & 25.43 & 0.933 & 0.875 & 0.739 & 0.838 \\
        EDFFDNet (ours) & \textcolor{blue}{30.63} & \textcolor{blue}{26.31} & \textcolor{blue}{22.15} & \textcolor{blue}{25.93} & \textcolor{blue}{0.938} & \textcolor{blue}{0.886} & \textcolor{blue}{0.763} & \textcolor{blue}{0.852} \\
		EDFFDNet-2 (ours) & \textcolor{red}{31.09} & \textcolor{red}{26.85} & \textcolor{red}{22.79} & \textcolor{red}{26.49} & \textcolor{red}{0.943} & \textcolor{red}{0.898} & \textcolor{red}{0.790} & \textcolor{red}{0.868} \\
		\bottomrule
        \vspace{-4mm}
	\end{tabular}
\end{table*}

\subsubsection{Correlation Computation}
In Table~\ref{tab:correlation}, we compare correlation methods for homography and FFD estimation. Global correlation in FFD reduces accuracy due to global field disturbances and increases inference time, while local correlation for homography struggles in low-overlap scenarios from insufficient search range. Our progressive strategy addresses these by providing adequate search range during global homography and finer range in refinement, improving accuracy by 0.39 dB PSNR and reducing inference time by 29.8\%.

\subsection{Comparative Experiments}
We conduct comprehensive comparative experiments for warping accuracy and computational overhead. In the experiments, we set the influence factor \(\theta = 0.75\), the number of groups \(N_{\mathrm{g}} = 8\), and the number of local refinement stages \(N_\mathrm{s}\) to either 1 or 2, corresponding to two versions: EDFFDNet and EDFFDNet-2. EDFFDNet uses \(M_1 = N_1 = 12\), while EDFFDNet-2 extends EDFFDNet with an additional local refinement stage where \(M_2 = N_2 = 18\).

\subsubsection{Comparison of Warping Accuracy}
We conduct a comprehensive comparison covering both traditional and deep learning-based approaches. The traditional methods included in our comparison are SIFT~\cite{lowe2004distinctive} + RANSAC~\cite{fischler1981random}, APAP~\cite{zaragoza2013projective}, ELA~\cite{li2017parallax},  SPW~\cite{liao2019single}, and LPC~\cite{jia2021leveraging}. For deep learning methods, we benchmark against UDIS~\cite{nie2021unsupervised}, MGDH~\cite{nie2021depth}, and UDIS++~\cite{nie2023parallax}.

\textbf{Quantitative Evaluation.} 
Table~\ref{tab:warp_quality_comparison} presents quantitative results on UDIS-D dataset~\cite{nie2021unsupervised}, where $I_{3\times3}$ denotes identity mapping. Metrics are categorized into three groups following prior works~\cite{nie2021unsupervised, nie2021depth, nie2023parallax}. For cases where traditional methods fail, we use identity mapping for evaluation. Our method outperforms previous approaches, especially in challenging scenarios, with significant improvements in hard and moderate categories, highlighting our model's enhanced locality.

\begin{figure*}[t]
  \centering
\vspace{-2mm}
   \includegraphics[width=0.9\linewidth]{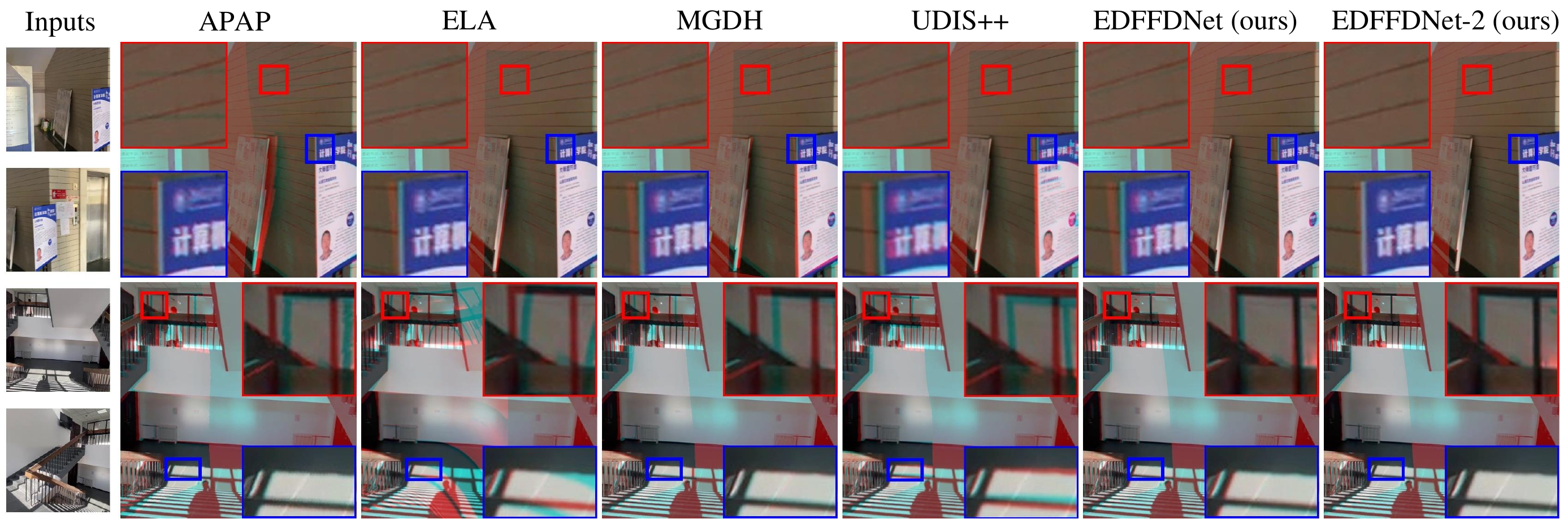}
\vspace{-3mm}
   \caption{Qualitative results on the UDIS-D dataset~\cite{nie2021unsupervised}. Zoomed-in regions from two distinct planes show multi-plane structures and depth disparities, highlighting the effectiveness of the proposed method in handling local deformations.}
   \label{fig:Qualitative comparison}
   \includegraphics[width=0.9\linewidth]{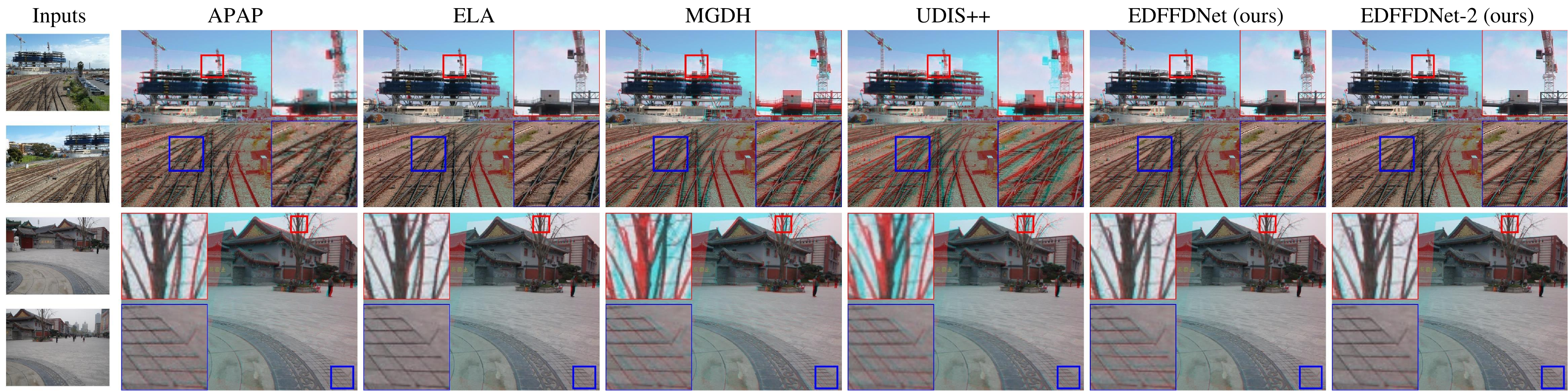}
\vspace{-3mm}          
   \caption{Qualitative results on the widely used cross-dataset cases: ``railtrack" (row-1) \cite{zaragoza2013projective} and ``temple" (row-2) \cite{gao2011constructing}.}
\vspace{-3mm}          
   \label{fig:Qualitative comparison2}
\end{figure*}

\textbf{Qualitative Evaluation.} 
Fig.~\ref{fig:Qualitative comparison} shows qualitative results on UDIS-D dataset \cite{nie2021unsupervised}, where we combine the green and blue channels of reference image $\mathbf{I}_{\mathrm{r}}$ with the red channel of result image $\mathbf{I}_{\mathrm{t}\rightarrow\mathrm{r}}$ for visual assessment. Traditional methods, MGDH, and UDIS++ achieve moderate alignment in hard scenarios with multiple planes and depth disparities requiring localized deformations. EDFFDNet demonstrates better performance through explicit locality, with EDFFDNet-2 achieving well-aligned results in challenging scenarios.

\textbf{Cross-dataset Evaluation.}
We further assess the generalization capability of our pre-trained model through cross-dataset validation. We first visualize widely used cases from \cite{zaragoza2013projective} and \cite{gao2011constructing} in Fig.~\ref{fig:Qualitative comparison2}, which demonstrates that our method, despite being trained only on the UDIS-D dataset~\cite{nie2021unsupervised} without cross-dataset fine-tuning, achieves superior generalization performance. We also evaluate zero-shot performance on the ScanNet~\cite{dai2017scannet} and ETH3D~\cite{schops2017multi} datasets in Table~\ref{tab:zero-shot}, which demonstrates the strong generalization capability of our method. Moreover, our method maintains competitive performance with traditional methods while requiring substantially less computational time, as shown in Table~\ref{tab:traditional_time}, highlighting its potential for practical deployment.

\begin{table}[H]
    \centering
    \footnotesize
    \renewcommand{\arraystretch}{1.2} 
\renewcommand\arraystretch{1}
    \setlength{\tabcolsep}{8pt} 
\vspace{-4mm}
    \caption{Zero-shot results on the testset of ScanNet and ETH3D. For ScanNet,10k test image pairs are randomly selected.}
\vspace{-3mm}
    \label{tab:zero-shot}
    \begin{tabular}{l c c c c}
        \toprule
        \multirow{2}{*}{ } & \multicolumn{2}{c}{ScanNet} & \multicolumn{2}{c}{ETH3D} \\
        \cmidrule(lr){2-3} \cmidrule(lr){4-5}
        & PSNR & SSIM & PSNR & SSIM \\
        \midrule
        UDIS~\cite{nie2021unsupervised} & 21.82 & 0.747 & 19.11 & 0.615 \\
        MGDH~\cite{nie2021depth} & 22.08 & 0.748 & 19.50 & 0.641 \\
        UDIS++~\cite{nie2023parallax} & 21.79 & 0.729 & 19.41 & 0.647 \\
        EDFFDNet (ours) & 23.37 & 0.786 & 20.54 & 0.693 \\
        EDFFDNet-2 (ours) & \textbf{24.32} & \textbf{0.808} & \textbf{21.47} & \textbf{0.731} \\
        \bottomrule
    \end{tabular}
\vspace{-7mm}
\end{table}

\begin{table}[H]
\renewcommand\arraystretch{1}
\renewcommand\tabcolsep{7pt}{}
\centering
\footnotesize
\vspace{-2mm}
\caption{Comparison of total time (s) required to generate the warping results.}
\vspace{-3mm}
\begin{tabular}{lccc}
\toprule
Dataset & Railtrack~\cite{zaragoza2013projective} & Fence \cite{lin2015adaptive} & Carpark \cite{gao2011constructing} \\ \hline
Resolution  & 1500 × 2000 & 1088 × 816  & 490 × 653  \\ \hline
APAP \cite{zaragoza2013projective}         & 159.583  & 55.560 & 23.743\\
ELA~\cite{li2017parallax}         & 19.251  & 8.867 & 5.760\\
SPW~\cite{liao2019single}          & 116.530 & 9.701 & 13.711\\
LPC~\cite{jia2021leveraging}           & 2114.944   & 11.913 & 68.947\\
EDFFDNet     & 0.078 & 0.055 & 0.053\\
EDFFDNet-2      & 0.097 & 0.064 & 0.063\\ 
\bottomrule
\vspace{-8mm}
\end{tabular}
\label{tab:traditional_time}
\end{table}

\subsubsection{Comparison of Computational Overhead}
We compare our method with deep learning approaches UDIS~\cite{nie2021unsupervised}, MGDH~\cite{nie2021depth}, and UDIS++~\cite{nie2023parallax} in terms of computational costs. Table~\ref{tab:overhead} shows EDFFDNet reduces parameters by 70.5\%, inference memory by 32.6\%, and total time by 33.7\% compared to UDIS++, while improving PSNR by 0.5 dB. With additional local refinement, EDFFDNet-2 achieves 1.06 dB PSNR improvement with low computational costs. These results highlight our method's accuracy and efficiency.

\begin{table}[h]
\renewcommand\arraystretch{1}
\renewcommand\tabcolsep{4pt}{}
\centering
\footnotesize
\vspace{-1mm}
\caption{Comparison of computational overhead.}
\vspace{-3mm}

\begin{tabular} {lccccc}
\toprule
        Method & PSNR & SSIM & \makecell{Parameters\\(M)} & \makecell{Memory\\(GB)} & \makecell{Total\\time (ms)} \\ \hline
        UDIS \cite{nie2021unsupervised} & 23.80 &  0.793 & 188.8  & 7.1 & 66.7  \\ 
        MGDH \cite{nie2021depth} & 24.89 & 0.817 & \textbf{16.4}  & 5.3 & 90.3  \\ 
        UDIS++ \cite{nie2023parallax} & 25.43 & 0.838 & 78.0  & 4.6 & 65.8  \\ 
        EDFFDNet (ours) & \underline{25.93} & \underline{0.852} & \underline{23.0}  & \textbf{3.1} & \textbf{43.6}  \\ 
        EDFFDNet-2 (ours) & \textbf{26.49} & \textbf{0.868} & 34.5  & \underline{4.3} & \underline{55.1}  \\
\bottomrule
\vspace{-10mm}
\end{tabular}
\label{tab:overhead}
\end{table}

\section{Conclusions}
We have proposed EDFFDNet, an unsupervised registration framework designed to handle large local deformations. This is achieved through an exponential-decay free-form deformation model for improved locality. Additionally, we have introduced an adaptive sparse motion aggregator that converts dense interactions into sparse ones for efficiency. We have also developed a progressive correlation strategy for coarse-to-fine estimation. Our approach delivers a significant advancement in both accuracy and efficiency.

\noindent\textbf{Acknowledgements.} This work was supported in part by the Zhejiang Provincial Natural Science Foundation of China under grant LD24F020003, in part by the National Natural Science Foundation of China under grant 62301484, in part by the Ningbo Natural Science Foundation of China under grant 2024J454, and in part by the National Key Research and Development Program of China under grant 2023YFB3209800.

{
    \small
    \bibliographystyle{ieeenat_fullname}
    \bibliography{arxiv}
}

\end{document}